\documentclass[letterpaper, 10 pt, conference]{ieeeconf}  

\IEEEoverridecommandlockouts                              
\usepackage[l2tabu,orthodox]{nag}

\usepackage[
    backend=bibtex8,
    style=ieee,
    sorting=none,
    natbib=true,
    doi=false,
    isbn=false,
    url=false,
    eprint=false,
    maxcitenames=1,
    mincitenames=1
]{biblatex}

\usepackage[pdftex,colorlinks]{hyperref}

\usepackage[printonlyused]{acronym}

\usepackage{siunitx}
\usepackage[all]{nowidow}

\usepackage[dvipsnames]{xcolor}

\usepackage{lipsum}

\usepackage{xspace} 
\newcommand{\ie}{i.e.,\xspace{}}

\usepackage[pdftex]{graphicx}

\usepackage{epstopdf}

\usepackage{import}

\graphicspath{{./latexGoodPractices/}}

\usepackage{booktabs}

\usepackage{tabularx}
\usepackage{multirow, multicol}

\usepackage{amssymb,amsfonts,amsmath,amscd}

\usepackage{bm}

\newcommand{\norm}[1]{\left\Vert#1\right\Vert}

\newcommand{\bbm}{\begin{bmatrix}}
\newcommand{\ebm}{\end{bmatrix}}

\usepackage{physics}  
\usepackage{amsmath} 

\makeatletter
\let\oldtheequation\theequation
\renewcommand\tagform@[1]{\maketag@@@{\ignorespaces#1\unskip\@@italiccorr}}
\renewcommand\theequation{(\oldtheequation)}
\makeatother

\title{\LARGE \bf 
    UAV-Assisted Self-Supervised Terrain Awareness \\for Off-Road Navigation
} 

\author{Jean-Michel Fortin$^{1}$, Olivier Gamache$^{1}$, William Fecteau$^{1}$, Effie Daum$^{1}$, William Larrivée-Hardy$^{1}$,  \\ François Pomerleau$^{1}$, Philippe Giguère$^{1}$
\thanks{*This research was supported by Fonds de Recherche du Québec Nature et technologies (FRQNT) Team grant 254912 and Natural Sciences and Engineering Research Council of Canada (NSERC) DRC Grant through the grant CRD 538321-18.}
\thanks{$^{1}$Northern Robotics Laboratory, Université Laval, Québec City, Québec, Canada
		{\texttt{\small jean-michel.fortin@norlab.ulaval.ca}} and \texttt{\small {philippe.giguere@ift.ulaval.ca}}}%
}

\usepackage[switch]{lineno}
\usepackage[font=small]{caption}
\usepackage{subcaption}

\acrodef{SLAM}{Simultaneous Localization And Mapping}
\acrodef{UGV}{Uncrewed Ground Vehicles}
\acrodef{UAV}{Uncrewed Aerial Vehicle}
\acrodef{FPV}{First-Person View}
\acrodef{LAGR}{Learning Applied to Ground Vehicles}
\acrodef{MCR}{Motor Currents versus Rate of turn}
\acrodef{ICR}{Instantaneous Centre of Rotation}
\acrodef{IMU}{Inertial Measurement Unit}
\acrodef{NMHE}{Nonlinear Moving Horizon Estimator}
\acrodef{CNN}{Convolutional Neural Networks}
\acrodef{TnR}{Teach-and-Repeat}
\acrodef{fps}{frames per second}
\acrodef{GNSS}{Global Navigation Satellite System}
\acrodef{PTP}{Precision Time Protocol}
\acrodef{BEV}{Bird's-Eye View}
\acrodef{PoV}{Point-of-View}
\acrodef{FoV}{Field-of-View}
\acrodef{PSD}{Power Spectral Density}
\acrodef{GSD}{Ground Sampling Distance}
\acrodef{RTK}{Real-Time Kinematic}
\acrodef{AUC}{Area Under the Curve}
\acrodef{POI}{Points of Interest}
\acrodef{MLP}{Multi-Layer Perceptron}
\acrodef{MPC}{Model Predictive Control}
\acrodef{RL}{Reinforcement Learning}
\acrodef{FFT}{Fast Fourier Transform}
\acrodef{RMSE}{Root Mean Squared Error}

\newcommand{\SE}{\mathrm{\mathbf{SE}}}

\begin{document}
\maketitle
\thispagestyle{empty}
\pagestyle{empty}

\begin{abstract}
Terrain awareness is an essential milestone to enable truly autonomous off-road navigation.
Accurately predicting terrain characteristics allows optimizing a vehicle's path against potential hazards. 
Recent methods use deep neural networks to predict traversability-related terrain properties in a self-supervised manner, relying on proprioception as a training signal.
However, onboard cameras are inherently limited by their point-of-view relative to the ground, suffering from occlusions and vanishing pixel density with distance.
This paper introduces a novel approach for self-supervised terrain characterization using an aerial perspective from a hovering drone.
We capture terrain-aligned images while sampling the environment with a ground vehicle, effectively training a simple predictor for vibrations, bumpiness, and energy consumption.
Our dataset includes \SI[detect-weight=true,mode=text]{2.8}{\kilo\meter} of off-road data collected in forest environment, comprising \num[detect-weight=true,mode=text]{13484} ground-based images and \num[detect-weight=true,mode=text]{12935} aerial images.
Our findings show that drone imagery improves terrain property prediction by \SI[detect-weight=true,mode=text]{21.37}{\percent} on the whole dataset and \SI[detect-weight=true,mode=text]{37.35}{\percent} in high vegetation, compared to ground robot images.
We conduct ablation studies to identify the main causes of these performance improvements.
We also demonstrate the real-world applicability of our approach by scouting an unseen area with a drone, planning and executing an optimized path on the ground.


\end{abstract}


\acresetall 

\section{Introduction}
\label{sec:introduction}


\ac{UGV} operating off-road should be capable of establishing a feasible path to their objective, while limiting mechanical wear and power consumption. 
This can only be accomplished through terrain awareness, which is a critical yet unsolved aspect of field robotics, especially in applications like forestry, search and rescue and planetary exploration~\citep{borges2022survey}.
In contrast to urban settings, off-road terrains are characterized by their heterogeneity and unpredictability.
The complexity of vehicle-terrain interactions arises from a multitude of factors. 
Soil characteristics such as geometry, density, cohesion, shear strength, and water content interact with vehicle properties including weight, locomotion method, geometry, and speed~\citep{wong2009terramechanics}.
This plethora of variables leads researchers to seek simplified environment representations to estimate the traversability of the terrain.

Early research on the topic focused primarily on geometric approaches, \ie{} leveraging range sensors to derive quantitative metrics such as slope and surface roughness~\citep{fankhauser2018probabilistic}.
However, these methods exhibit limitations in environments where geometry data alone is insufficient to determine traversability, such as in the presence of mudholes, tall vegetation, sand or ice patches~\citep{borges2022survey}. 
Alternatively, appearance-based methods employ deep neural networks for terrain classification~\citep{viswanath2021offseg}, but exhibit significant challenges in generalizing to unobserved classes or changes in environmental conditions~\citep{guastella2020learning}.
Furthermore, classification is constrained by inherent variability, as a patch of dirt will fluctuate significantly in terms of cohesion and shear resistance based on its moisture content~\citep{wong2009terramechanics}.

\begin{figure}[t]
	\centering
	\includegraphics[width=0.48\textwidth]{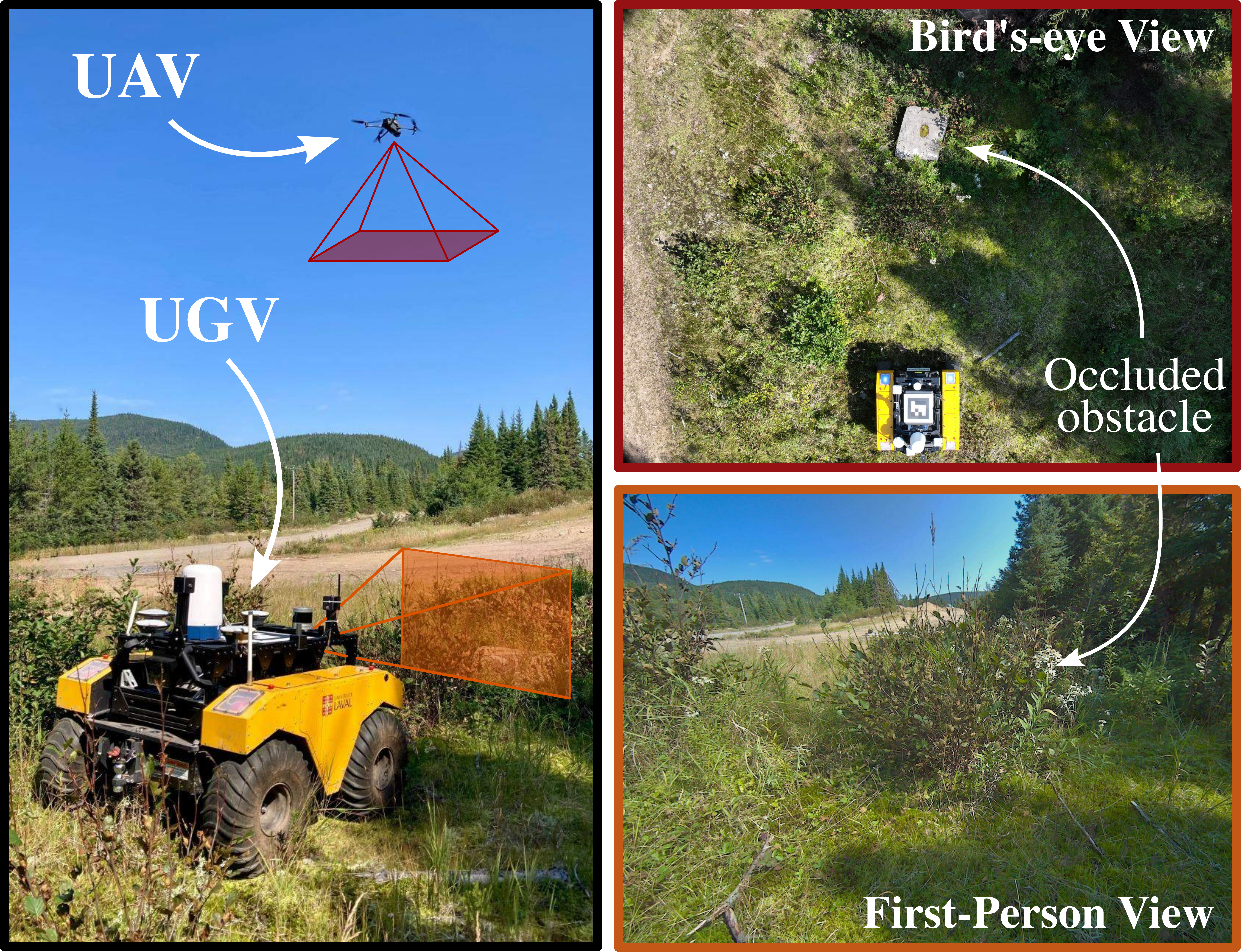}
	\caption{Visual representation of our experimental setup. We propose a novel approach for self-supervised terrain awareness for \acf{UGV}, leveraging an aerial viewpoint from a hovering \acf{UAV}. As illustrated, the \acs{UAV} offers a perspective that is better aligned with the terrain and experiences fewer occlusions.} 
	\label{fig:intro}
        \vspace{-0.15in}
\end{figure}

Self-supervised learning has recently gained traction as a promising avenue to enhance the reliability of terrain characterization \citep{borges2022survey, guastella2020learning}. 
This approach enables robots to learn from their own experience, effectively reducing the need for manual labeling, accounting for vehicle capabilities, and opening the possibility of lifelong learning.
However, the prediction of terrain properties from a distance is limited by the selection of exteroceptive sensors, particularly in terms of their \ac{PoV} relative to the ground.
In fact, a camera or lidar placed on a mobile robot can easily miss a ditch or be obstructed by tall grass, as highlighted in \autoref{fig:intro}.
Moreover, the pixel density on the ground decreases quadratically with distance, leading to imprecise predictions from afar. 

Consequently, we propose a novel approach for self-supervised terrain characterization that leverages an aerial perspective. 
While \ac{UGV}s can suffer momentary pitch and roll motions, the outdoor world is a predominantly two-dimensional environment.
In this scenario, the addition of a top-down perspective of the surroundings holds significant potential.
\autoref{fig:intro} provides a visual representation of this configuration, illustrating an \ac{UAV} positioned above a \ac{UGV}, offering an elevated point that facilitates predictive analysis of the terrain and potential obstacles along the vehicle path.
To the best of our knowledge, this is the first work on self-supervised terrain characterization that compares the terrain analysis capabilities of aerial against ground-based images.
In short, our contributions are the following:
\begin{enumerate}                                             
    \item A novel approach for self-supervised terrain prediction, leveraging an aerial viewpoint;
    \item An analysis of the improvements of using aerial imagery over ground-based cameras; and
    \item A real-world demonstration of the potential of aerial scouting to plan missions ahead for \acp{UGV}.
\end{enumerate}

\section{Related Work}
\label{sec:related_work}

\subsection{Estimating Terrain Properties In Situ} 


Terramechanics is the study of the interaction between a vehicle and its underlying terrain \citep{wong2009terramechanics}.
Whereas studies of soil properties usually require dedicated equipment, our objective is to estimate terrain properties \textit{in situ} -- through direct ground contact -- using standard onboard robot sensors.
Early research in this field explored different sensor modalities for terrain classification.
\citet{ojeda2006terrain} highlighted that the roll and pitch velocities, the acceleration on the Z-axis, along with power consumption, were particularly effective in distinguishing between gravel, grass, sand, pavement and dirt.
Building on these conclusions, \citet{reina2016slip} developed methods to predict specific terrain properties such as motion resistance coefficients and slip track -- the distance of the instantaneous center of rotation along the Y-axis -- for skid-steering vehicles. 
Both approaches use data from an \ac{IMU}, wheel encoders, and motor current sensors.
More recently, deep neural networks have been employed to analyze spectrograms of combined inertial and motor data for terrain classification~\cite{vulpi2021recurrent,larocque2024proprioception}.

These studies offer valuable insights into the use of standard onboard sensors and robot-terrain interactions to estimate terrain properties.
However, they are primarily limited to \textit{in situ} predictions, which is unsuitable for path planning or predictive control. 
In the context of this research, we are interested in leveraging proprioceptive sensing to predict terrain properties \textit{from a distance}.



\subsection{Predicting Terrain From a Distance}

Predicting terrain properties in advance requires the use of exteroceptive sensors.
High-dimensional or unstructured data sources, such as point clouds or images, introduce significant challenges, leading researchers to exploit the strengths of deep neural networks.
These networks excel at learning features in high-dimensional spaces that correlate with specific tasks.
Over the last decade, various approaches have been explored to predict traversability, which refers to a ground vehicle's ability to navigate a particular terrain.
Although geometry analysis \citep{fankhauser2018probabilistic} and terrain classification \citep{viswanath2021offseg} have shown promise in enhancing traversable path planning for off-road conditions, they face limitations in their ability to identify non-geometric hazards and capture the full spectrum of terrain properties.
Consequently, our analysis emphasizes on self-supervised terrain assessment methods, which promise to be more scalable, platform-aware and adaptive to new environments. 




Determining the appropriate training signal and its acquisition method is complex and largely depends on the robot's intended task.
To maximize passenger comfort, \citet{castro2023does} devised vibration metrics that correlate with the Z-axis acceleration of the vehicle, while \citet{yao2022rca} also included roll and pitch velocities, all measured by an \ac{IMU}. 
Some researchers analyze the discrepancies between predicted and actual motion to derive traversability scores~\citep{frey2023fast} or to train a terrain-aware motion models~\citep{maheshwari2023piaug, gasparino2022wayfast}.
In the case of legged robots, force-torque sensors mounted on the robot's feet can extract ground reaction scores through wavelet analysis~\citep{wellhausen2019should}.
\citet{karnan2023sterling} demonstrate a self-supervision method to learn visual embeddings closely related to raw proprioceptive measurements, without any manual data processing.
Lastly, carefully tuned traversability heuristics derived from geometric analysis can serve as training signals \citep{frey2024roadrunner, chen2023learning}, acknowledging that the quality of the predictor is upper bounded by these heuristics.
While all these approaches have merit, our work aims to demonstrate that terrain properties can be predicted more accurately from top-down images. 
Therefore, we focus on predicting vibrations and power consumption, as proxies that correlate more closely with robot-terrain interactions.


The aforementioned works primarily use cameras, sometimes combined with depth sensors, as input sources. 
These cameras are typically mounted on the \ac{UGV}, offering a limited perspective of the surrounding terrain.
While many researchers work directly with \ac{FPV} images, some have demonstrated the advantages of projecting these images to a \ac{BEV} for improved coherence and easier sensor fusion~\citep{castro2023does, yao2022rca, maheshwari2023piaug, karnan2023sterling}.
However, the \ac{BEV} approach has its limitations. 
As the distance from the camera increases, pixel density experiences a quadratic decay, resulting in blurry images, especially when projected over a depth map~\citep{castro2023does}. 
Moreover, \ac{FPV} image quality significantly deteriorates when the UGV experiences high pitch and roll motions, frequent in off-road driving, further complicating the projection process.
To address these challenges, we propose to leverage aerial imagery from a hovering drone to predict terrain properties from a consistent, high-quality overhead view of the terrain. 

\subsection{UGV-UAV Collaboration}

With the growing popularity of multi-robot systems and swarm robotics, there has been a push to explore how to combine sensors positioned on different robots.
For instance, in search and rescue scenarios \citep{stentz2003real, delmerico2017active}, \acp{UAV} are used for scouting or mapping the environment before deploying the \ac{UGV}.
Early experiments by \citet{stentz2003real} demonstrated that including aerial information yields better global planning and safer trajectories for the ground vehicle, especially in the presence of negative obstacles, such as holes or ditches.
More recently, \citet{delmerico2017active} implemented a similar system, in which a drone equipped with a monocular camera creates a grid of the environments with terrain classification and elevation data for safe path planning on the ground.
However, experiments were limited to urban areas with distinct terrain types. 
Recent work achieved effective collaboration by using \acp{UAV} equipped with downward-facing lidars and \acp{UGV} to generate large-scale cross-view risk maps~\citep{wang2024risk}.
Furthermore, the authors exploit the semantic information as a predictor of terrain traversability, enabling \acp{UGV} to assess unvisited areas using aerial maps and expanding their operational awareness.

Finally, \citet{manderson2020learning} demonstrated the potential of combining aerial and onboard camera images by training an off-policy \ac{RL} agent to avoid rough terrain.
Sparse aerial images are fed as additional observations to the agent, which resulted in a consistent improvement in navigating over smooth terrain compared to relying solely on a \ac{FPV}.
While these results are very encouraging, our study focuses on analyzing the limitations of the onboard viewpoint and the advantages of the aerial one.  
We combine the scalability of self-supervised learning with the flexibility, increased \ac{FoV} and uniform spatial resolution of \acp{UAV} to predict terrain properties more accurately and explore unknown areas beforehand.


\section{Methodology}
\label{sec:methodology}


\begin{figure*}[htbp]
    \centering
    \includegraphics[width=0.97\textwidth]{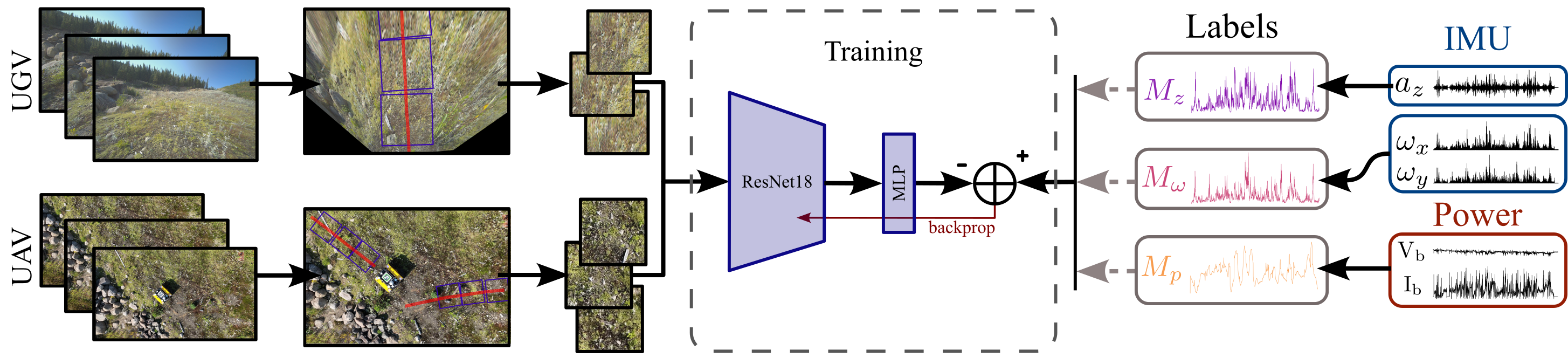}
    \caption{Overview of our pipeline for data collection and terrain properties learning. 
    (Top Left) Extraction of image patches along the robot's path from onboard UGV images involves projecting the image into a \ac{BEV} and extracting patches at specified points of interest.
    (Bottom Left) For aerial images, the same patches are extracted relative to the robot's position, identified by an \textit{Aruco} marker.
    (Middle) Depiction of the terrain predictor, featuring a ResNet architecture coupled with a fully-connected network for the regression of a single terrain-related metric.
    (Left) Application of proprioceptive measurements to generate labels that correlate with terrain characteristics.
    }
    \label{fig:pipeline}
    \vspace{-0.15in}
\end{figure*}

\subsection{Data Collection}

We used a half-ton \textit{Clearpath Warthog} attached with an \textit{Aruco} marker on top to facilitate visual detection. 
Data collection was performed at the \textit{Montmorency} research forest, in areas where unobstructed aerial views were possible.
Selected terrains included a large quarry, forest trails, high vegetation, rough gravel, and a mossy area. 
The assembled dataset contains camera images from a \textit{ZED X} stereo camera (\SI{5}{\hertz}), accelerometer and gyroscope measurements from an \textit{Xsens Mti-30} \ac{IMU} (\SI{100}{\hertz}), power measurements from the main battery (\SI{10}{\hertz}) and \ac{GNSS} measurements from an \textit{Emlid Reach M2} receiver (\SI{5}{\hertz}), with \ac{RTK} corrections from a base station.
Simultaneously, a \textit{DJI Mavic 3E} drone hovered approximately \SI{10}{\meter} above the robot, recording 4K videos at \si{30} \ac{fps}.
Time synchronization inside the \ac{UGV} is assured using the \ac{PTP} over the Ethernet network, and drone images were synchronized manually by finding the video frame where the vehicle initiates movement.

The robot was set to drive forward at a constant velocity of \SI{1.5}{\meter \per \second}, which is a compromise between discerning ground vibrations and limiting the vehicle's natural oscillations at high speed. 
A human supervisor was present to alter the robot's  trajectory to prevent collisions or avoid moving out of the drone's sight.
Overall, the collected dataset accounts for \SI{45}{minutes} of driving, for a total of \si{7536} unique terrain patches over a distance of \SI{2.8}{km}. 
The ground dataset is composed of \num{13484} images and the aerial dataset contains \num{12935} images.


\subsection{Patch Extraction}
\label{sec:patch_extraction}

Our predictor is trained using terrain patches extracted along the robot's traveled path. 
This pipeline, displayed in the left part of \autoref{fig:pipeline}, first requires projecting the \ac{UGV}'s front camera images to a \ac{BEV} format.
We make the hypothesis that the ground  is locally flat around the robot, and then find the rigid transformation ${}^G_C\mathbf{T} \in \SE(3)$ between the ground plane $G$ and the camera frame $C$.
However, as the robot suffers from pitch and roll motion during data acquisition, this transformation cannot be assumed to be fixed. 
Our calibration procedure determines the extrinsic transformation ${}^R_C\mathbf{T}$ between the camera's optical frame and the robot's base link $R$ by placing a chessboard on a flat ground in front of the robot.
Then, we can estimate the transformation ${}^G_R\mathbf{T}$ between the robot frame and the ground plane during motion, using the roll and pitch calculated by the \ac{IMU}'s Madgwick filter, resulting in ${}^C_G\mathbf{T} = {}^C_R\mathbf{T} \: {}^R_G\mathbf{T}$.

For aerial images, we maintain the locally flat ground assumption, and consider that drone images are rectified and normal to the ground plane, which is assured by the \ac{UAV}'s gimbal compensation. 
This yields a constant \ac{GSD} in the image, computed by dividing the perimeter of the marker (in pixels) by its known perimeter (\SI{1.4}{m}).
Similarly, the robot's heading relative to the image plane is aligned by taking the vector going through the marker's center.

Terrain patches are sampled at specific points of interest along the traversed path, which are the same for both onboard and aerial images.
Wheel odometry measurements are used to filter out timestamps where the robot was not going straight, effectively removing manually driven segments.
Points of interest are taken from the \ac{RTK} localization at fixed distance intervals of \SI{30}{\centi\meter} to distribute data evenly.  
The extracted patches measure \SI{1.5}{\meter} x \SI{1.5}{\meter}, which is roughly the robot's footprint on the ground.
The pose of the robot for each image acquisition is interpolated linearly between localization timestamps.
As the same terrain patch is seen more than once, the multiple views are accumulated in a common folder and one patch is randomly sampled from it during training, increasing viewpoint invariance implicitly \citep{karnan2023sterling}. 


\subsection{Label Generation}
\label{sec:label_generation}

Self-supervised terrain prediction requires a training signal that correlates with the robot's interaction with the terrain.
In this work, we define three terrain-related metrics to evaluate the capabilities of our trained models: a vibration metric $M_z$, a bumpiness metric $M_{\omega}$, and a power consumption metric $M_{p}$.
We derive $M_z$ by analyzing the frequency spectrum of the \ac{IMU}-measured Z-axis acceleration $a_z$, as in \citep{castro2023does}. 
Specifically, we compute the power spectral density $S^{W}_{a_z}$ over a time window $W$ using Welch's method \citep{welch1967use} and calculate the bandpower $B$ as
\begin{equation}
    \mathrm{B}(t) = \int_{f_{min}}^{f_{max}} S^W_{a_z}(f) \, df,
\label{eq:traversability-cost}
\end{equation}
where $f_{min}=\SI{1}{\hertz}$ and $f_{max}=\SI{30}{\hertz}$.
The evaluation window $W$ represents the time required for the robot to traverse a terrain patch equal to its own length.
We refer to this time as $2\alpha$, meaning that a window is given by $W = [t - \alpha, t + \alpha]$.
To mitigate the influence of extreme peaks and achieve a more balanced distribution of labels for training, we apply a logarithmic transformation to $B$. 
Additionally, we shift the resulting distribution to ensure all values remain positive.
Using \ref{eq:traversability-cost}, the final vibration metric is calculated using
\begin{equation}
    M_z(t) = \ln\left(1 + \mathrm{B}(t)\right).
\label{eq:zacc_metric}
\end{equation}
Inspired by the findings of \citet{ojeda2006terrain}, we use the roll and pitch angular velocities ($\omega_x, \omega_y$) to devise a bumpiness metric $M_\omega$ as
\begin{equation}
    M_\omega(t) = \norm{
    \begin{bmatrix}
           \omega_x(t) \\
           \omega_y(t) \\
    \end{bmatrix}
    }.
\label{eq:omega_metric}
\end{equation}
To predict power consumption, we introduce a metric $M_{p}$ that quantifies the electrical energy required for the vehicle to traverse a terrain patch equal to its own length.
This metric is calculated using the following formula:
\begin{equation}
    M_{p}(t) = \sum_{t - \alpha}^{t + \alpha} I_b(t) V_b(t) \Delta t,
\label{eq:power_metric}
\end{equation}
where $I_b(t)$ and $V_b(t)$ represent the measured current and voltage of the main battery at time $t$, respectively, while $\alpha$ is defined as previously mentioned. 
A discretized Gaussian filter of parameters $\mathcal{N}(0, 0.5^2)$ was applied on each of the presented metrics, making the analyzed signals smoother.
A sample of the raw measurements and generated metrics is presented on the right part of \autoref{fig:pipeline}.


\subsection{Terrain Predictor}

We train a deep neural network from the previously described data to predict terrain properties from images of terrain patches.
Since \ac{CNN} perform well on texture analysis, we selected a ResNet18 backbone \citep{he2016deep}, followed by a three-layer fully-connected network, to regress a single value.
This network is represented in the center of \autoref{fig:pipeline}.

During inference, we use a sliding window to extract a grid of terrain patches from the \ac{BEV} images.
These patches overlap on each other to smooth the resulting predictions and mitigate the effect of false negatives. 
Local maps of predicted properties are generated by passing this batch of patches through the trained predictor, and reconstruct the final map by averaging the predictions for each pixel. 


\section{Experiments and Results}
\label{sec:results}

We hereby present our experimental protocol and results, focusing on the benefits of aerial perspective in terrain analysis. 
We explore whether an aerial \ac{PoV} enhances our understanding of specific terrain properties and investigate the factors that influence prediction accuracy. 
In addition, we demonstrate the scalability of our approach for real-world deployments. 


\subsection{Implementation Details}

The predictor was trained on an \emph{Nvidia RTX-3090} GPU, with a batch size of \num{256} and a learning rate of \num{5e-5}. 
Input patches resized to \num{256} x \num{256} pixels, and applied min-max normalization. 
Each conducted experiment follows a 5-fold cross-validation procedure to mitigate the effect of the small dataset. 
The training set contains \SI{90}{\percent} on the data samples, whereas the validation set holds the remaining \SI{10}{\percent}.
Output labels correspond to one of the previously defined terrain-related metrics, with min-max applied normalization over the whole training set.
Consequently, the predictor's output is passed through a sigmoid activation function to predict normalized values in $[0, 1]$.
For the real-world experiment, this network was deployed on an \emph{Nvidia Jetson Orin AGX} embedded computer. 

\subsection{Aerial images for terrain prediction}
\label{sec:results_training}


Our initial hypothesis is that a top-down view of the terrain provides better insight on its properties than an onboard camera with an oblique viewpoint.
To validate this hypothesis, we perform some experiments on the assembled patch dataset. 
\autoref{tab:training_results} shows the average \ac{RMSE} of the prediction on a subset of data, using 5-fold cross-validation. 
Note that for \ac{UGV} patch extraction, we kept only patches that were \SI{5}{\meter} or closer to the camera, to reduce the impact of projection blur and provide a fair comparison.
The results highlight that the predictor relying on aerial patches performs on average \SI{21.37}{\percent} better on all metrics and subsets of data than the one relying on ground-based ones. 
Interestingly, we notice that the discrepancy is much higher on the high vegetation dataset, with an improvement margin of \SI{37.35}{\percent}.
This can be explained by the obstructed \ac{PoV} of the \ac{UGV}, which cannot distinguish what is behind the grass, as illustrated in \autoref{fig:intro}. 

\begin{figure}[tbp]
	\centering
	\includegraphics[width=0.48\textwidth]{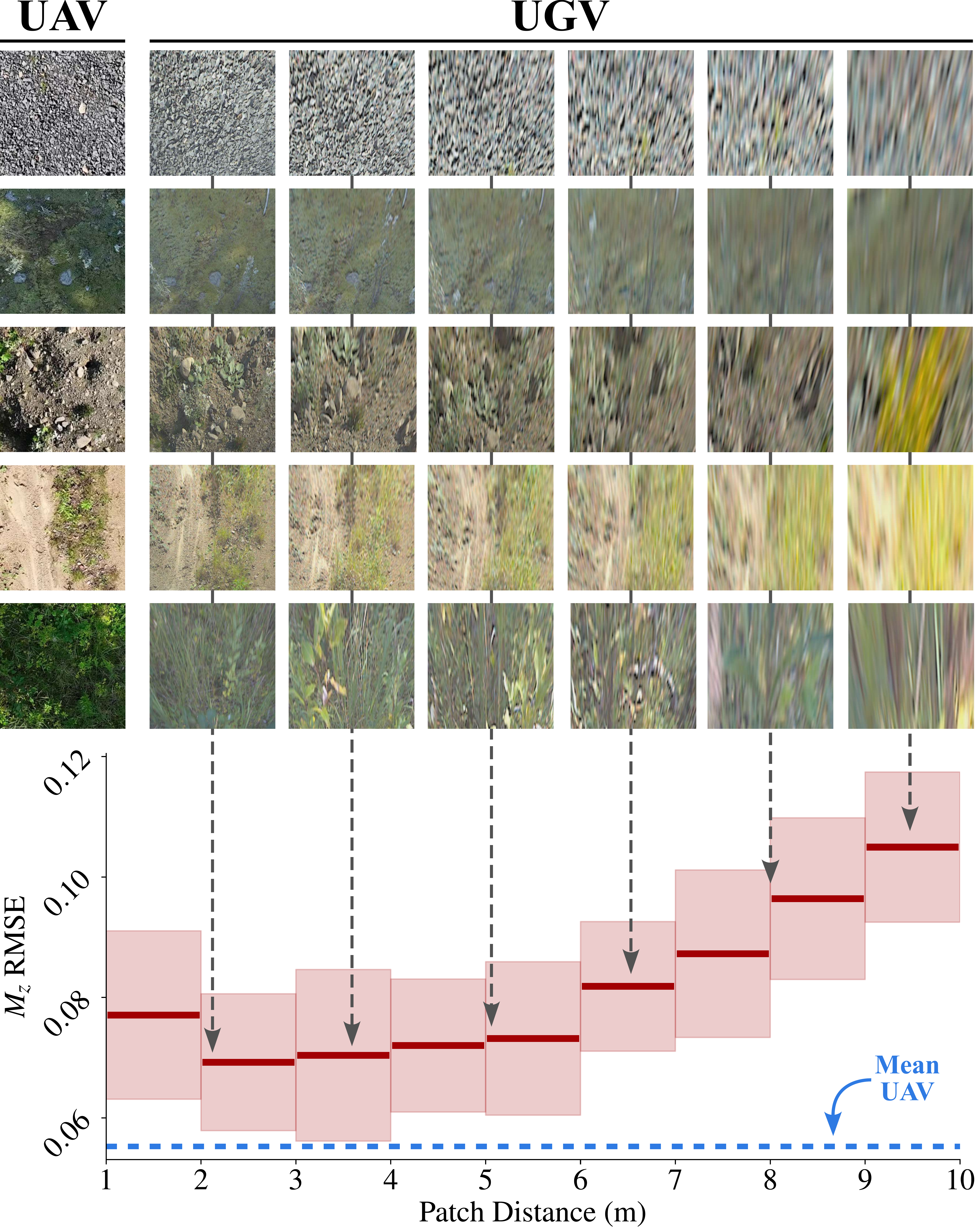}
	\caption{Ablation study showing the impact of patches' distance to the \ac{UGV}'s camera on RMSE of prediction of $M_z$. The quantitative results are complemented with typical ground images at different distances, where each row is for a different terrain patch seen from below by the vehicle. The same patch acquired from the \ac{UAV} is shown on the left, highlighting the absence of deformation compared to ground images.} 
	\label{fig:impact_distance}
        \vspace{-0.1in}
\end{figure}

\begin{table}[tbp]
    \def\arraystretch{1.3}
    \setlength{\tabcolsep}{5.5pt}
    \centering
    \begin{tabular}{@{}lcc|cc|cc@{}}
         \hline
         \multirow{2}{4em}{Terrain} & \multicolumn{2}{c}{Vibration ($M_{z}$)} & \multicolumn{2}{c}{Angular vel.  ($M_{\omega}$)} & \multicolumn{2}{c}{Power ($M_{p}$)} \\ \cline{2-7}
          & UGV & UAV & UGV & UAV & UGV & UAV \\ 
         \hline 
    Quarry & 0.0911 & \underline{0.0671} & 0.0693 & \underline{0.0600} & 0.0510 & \underline{0.0374} \\
    Trails & 0.0819 & \underline{0.0755} & 0.0742 & \underline{0.0600} & 0.0346 & \underline{0.0283} \\
    Mossy & 0.0825 & \underline{0.0447} & 0.0490 & \underline{0.0436} & 0.0361 & \underline{0.0245} \\
    Gravel & 0.0906 & \underline{0.0686} & 0.0616 & \underline{0.0539} & 0.0361 & \underline{0.0316} \\
    Vegetation & 0.1389 & \underline{0.0806} & 0.1217 & \underline{0.0794} & 0.1285 & \underline{0.0831} \\
    \hline
    \textbf{Whole} & 0.0728 & \textbf{0.0539} & 0.0592 & \textbf{0.0458} & 0.0374 & \textbf{0.0316} \\
    \hline
    \end{tabular}
    \caption{Prediction RMSE on the assembled aerial (UAV) and ground (UGV) datasets, for the three metrics. Shown results are averaged using a 5-fold cross-validation procedure.} 
    \label{tab:training_results}
\end{table}
Another factor of high influence on the performance of the predictor on ground-based images is the reduction of the pixel density with distance.
Indeed, as shown in the top part of \autoref{fig:impact_distance}, the quality of extracted terrain patches decreases drastically with their distance from the camera. 
Without loss of generality, for a camera with its optical axis parallel to the ground, the footprint of a pixel projected on the ground, and seen from \ac{BEV}, grows to the square of the distance.
Specifically, the height of one pixel on the ground is equal to \SI{2.34}{\milli\meter} for a patch located at \SI{1}{\meter} from the robot, compared \SI{14.38}{\centi\meter} at \SI{10}{\meter}.
In comparison, the drone images provide an evenly distributed \ac{GSD} of \SI{3.66}{\milli\meter} when capturing images at \SI{10}{\meter} of altitude. 
The bottom graph in \autoref{fig:impact_distance} shows the results of an experiment where the predictor was trained with all the extracted terrain patches up to a distance of \SI{10}{\meter} from the robot.
The trained model was evaluated on terrain patches located further and further away from the robot, in intervals of \SI{1}{\meter}. 
The results validate that performances decrease
monotonically with the distance from the camera, which correlates with the decrease in spatial resolution. 
The first step is an outlier and stems from the fact that the dataset contains fewer patches located between 1 and 2 meters, considering the camera angle and the patch width. 
Notice how the mean prediction error for the aerial dataset is still below the lowest value of the ground dataset, confirming that others factors, such as occlusions and viewing angle, also affect the predictions. 
\begin{figure}[!h]
	\centering
	\includegraphics[width=0.48\textwidth]{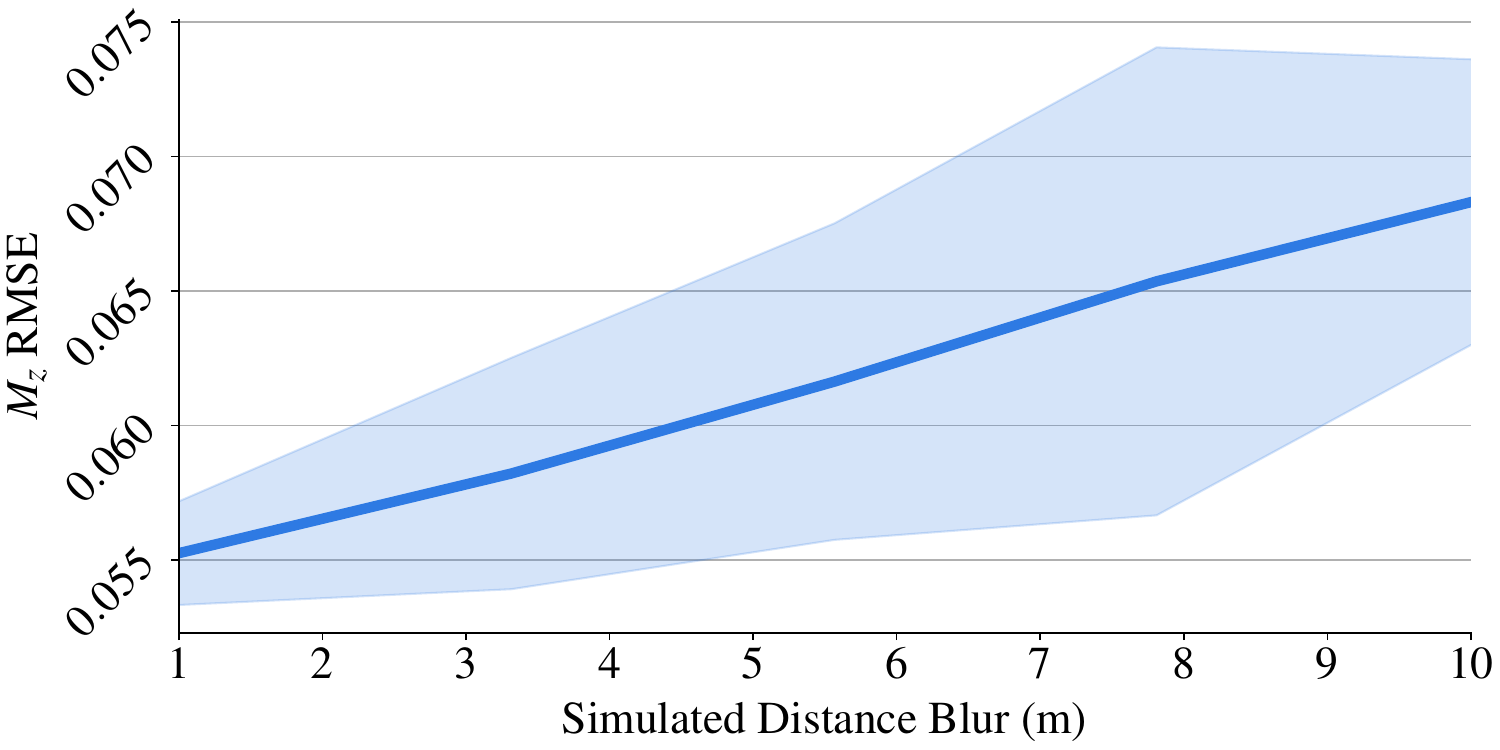}
	\caption{Impact of blurring aerial patches on RMSE of prediction of $M_z$. The X-axis is a simulated distance based on the \ac{GSD} increasing quadratically with the distance. To simulate the observed effects on ground images, we increased the blurring kernel's standard deviation in the Y-axis.}
	\label{fig:impact_blur}
        \vspace{-0.15in}
\end{figure}
To further prove this trend, we conduct a second test, where we artificially blurred the aerial patches with a dominantly vertical Gaussian kernel. 
We adjusted the kernel's vertical standard deviation to the equivalent pixel size for ground patches, which grows to the square of the horizontal distance between the camera and the patch.  
The network was trained multiple times with increasing blur distributions and evaluated on images with the same blurring function applied. 
The resulting RMSE curve is displayed in \autoref{fig:impact_blur}, and highlights a similar trend than the distance experiment in \autoref{fig:impact_distance}, confirming that projection blur is one of the main causes limiting the \ac{UGV}'s prediction accuracy.

The previous experiments lead us to conclude that the prediction range of the \ac{UGV} is constrained to a maximum \SI{6}{\meter} due to the quadratic growth of the blur with distance.
Elevating the camera position would not adequately address this issue, as the ground area represented by each pixel only expands linearly with camera height.
This limitation raises significant doubts about the effectiveness of any method relying on onboard cameras for high-speed terrain assessment, even before factoring in the additional complications of motion blur.


\subsection{Real-World Demonstration}

\begin{figure}[htbp]
    \vspace{0.08in}
	\centering
	\includegraphics[width=0.48\textwidth]{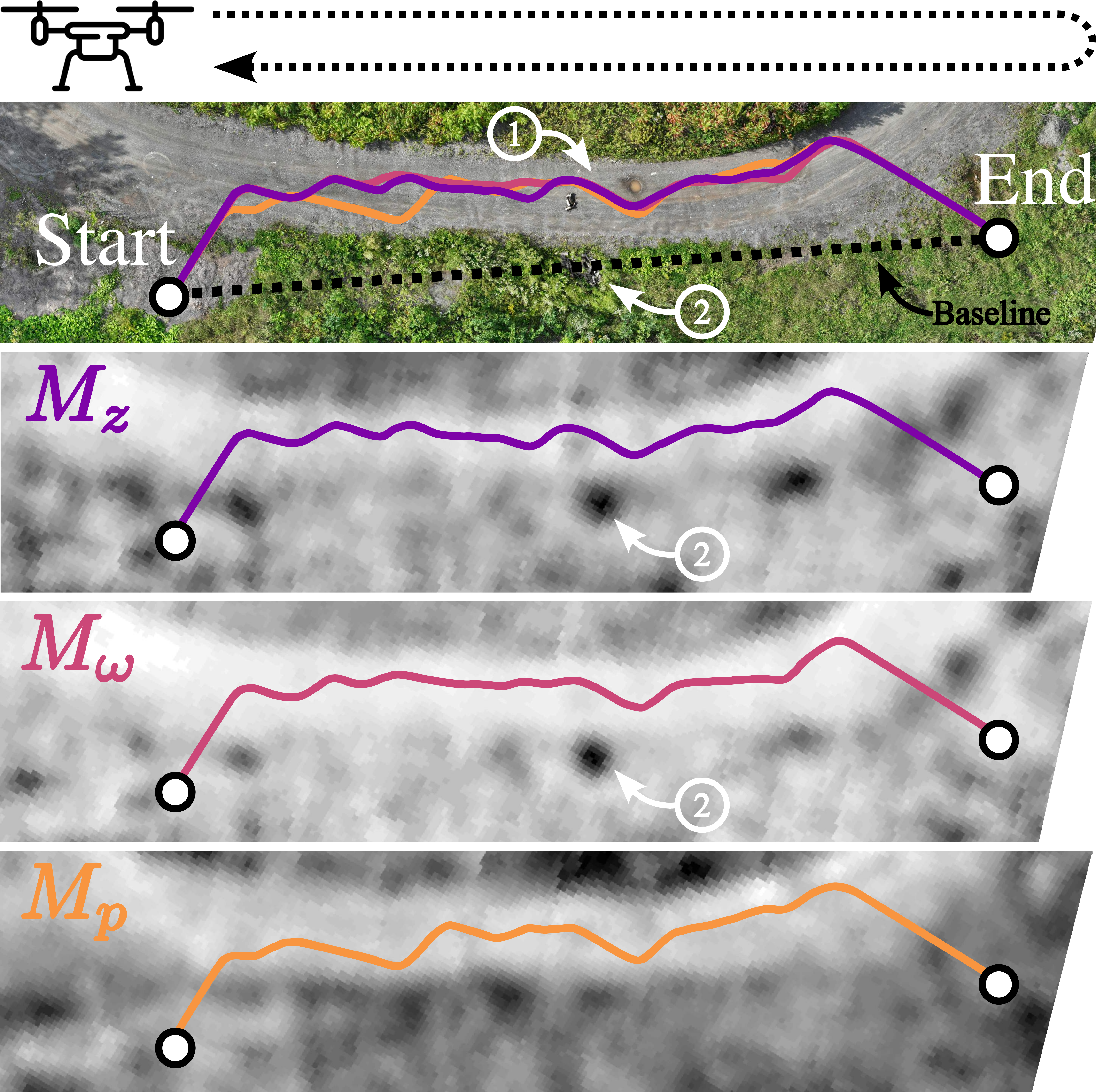}
	\caption{Field demonstration highlighting our method capability to efficiently estimate the ground properties from a drone survey. 
 A terrain map and an optimized trajectory is calculated for each developed metric. The baseline is the shortest feasible path from the start point to the end point. 
 \raisebox{.5pt}{\textcircled{\raisebox{-.9pt} {1}}} Two obstacles were avoided by all three metrics.
 \raisebox{.5pt}{\textcircled{\raisebox{-.9pt} {2}}} Detection of a large hole by $M_z$ and $M_{\omega}$.
 }
	\label{fig:real_world_experiment}
        \vspace{-0.15in}
\end{figure}

The ability to deploy a \ac{UAV} above an unknown terrain allows for faster scouting and assessment of the ground properties without risking \ac{UGV} entrapment in unforeseen terrain obstacles.
Building on our previous experiments, we demonstrate the real-world applicability of our system by using \ac{UAV}-captured images of an unexplored environment to predict an optimized path for our \ac{UGV}.
As illustrated in the upper portion of \autoref{fig:real_world_experiment}, we deployed the \ac{UAV} on Université Laval's campus, an area not included in our training dataset, to survey a designated field.
Using the acquired images and \ac{RTK} \ac{GNSS} localization of the drone, we construct a local traversability map based on each of our three metrics.
\autoref{fig:real_world_experiment} displays these predicted terrain maps along with optimized paths, calculated using Dijkstra's shortest path algorithm.

By observing the resulting traversability maps, we note several elements that confirm the network's comprehension of the terrain.
Interestingly, all three metrics were able to segment the road from the vegetation, and identified the road as the preferred path for our \ac{UGV}.
This effect is particularly pronounced for $M_{p}$, since driving on vegetation is more power consuming than fine gravel roads.
As highlighted by the marker \raisebox{.5pt}{\textcircled{\raisebox{-.9pt} {1}}}, each metric successfully identified the small logpile and the puddle, and planned optimal trajectories that navigate between them.
Moreover, we notice that $M_z$ and $M_{\omega}$ confidently detected a large ground depression, marked by \raisebox{.5pt}{\textcircled{\raisebox{-.9pt} {2}}}.
Given that $M_{z}$ is based on \ac{UGV} vibrations, and $M_{\omega}$ on roll and pitch angular velocities, such obstacles impact greatly the estimated traversability cost. 

We executed two trajectories from \autoref{fig:real_world_experiment} with our \ac{UGV}, the baseline shown as a dashed black line and the $M_z$ optimized path in purple. 
The baseline consists of the shortest traversable path from the start point to the end, and resulted in the \ac{UGV} struggling to achieve its goal due to highly uneven terrain. 
On the contrary, the optimized path from $M_{z}$ showed the \ac{UGV} effectively avoiding environment obstacles which resulted in a smooth traversal of the unknown area.



\section{Conclusion}
\label{sec:conclusion}
In this paper, we proposed a novel method for self-supervised terrain characterization leveraging an aerial perspective. 
Our results show a significant improvement in terrain predictions, as the predictor trained on aerial images consistently perform \SI{21.37}{\percent} better compared to ground-based images, as evaluated on three distinct terrain-related metrics.
The disparity is more significant in high vegetation, exhibiting an improvement margin of \SI{37.35}{\percent}, attributed to important occlusion.
We analyze the reasons for this improvement through extensive ablation studies and qualitative analysis, focusing primarily on image deformation and limited pixel density caused by the \ac{PoV} of the \ac{UGV}'s camera.
We show that the prediction range of our \ac{UGV} is limited to \SI{6}{m}, raising serious questions about the value of \ac{FPV} images for off-road terrain assessment.
Finally, we integrated our system in a real-world scenario for aerial scouting of unseen areas and optimized path planning for a \ac{UGV}, and successfully traveled a path calculated from the drone images' predictions.

Looking ahead, we see several promising directions for further research.
Our current approach, which relies on top-down monocular images, lacks crucial geometric data. 
To address this, we propose incorporating photogrammetry techniques to generate depth information. 
This would allow providing an additional elevation layer to the predictor, significantly improving the detection of geometric obstacles and uneven terrain.
Additionally, we aim to tackle the challenge of snow characterization, which is particularly difficult to assess using visual data alone.
Hence, we plan to integrate new sensor modalities, such as thermal imaging, to provide a more comprehensive understanding of the terrain. 


\printbibliography

\end{document}